\documentclass[letterpaper, 10 pt, conference]{Packages/ieeeconf}  % Comment this line out if you need a4paper
\usepackage{microtype}  % 단어 간격 최적화
\microtypesetup{protrusion=true,expansion=true}
\IEEEoverridecommandlockouts                              % This command is only needed if ß
\usepackage{float}
\usepackage{times}
\usepackage{epsfig}
\usepackage{graphicx}
\usepackage{amsmath}
\usepackage{amssymb}
\usepackage{xcolor}
\usepackage[ruled]{algorithm}
\usepackage{algpseudocode}
\usepackage{ctable}
\usepackage[export]{adjustbox}
\usepackage{import}
\usepackage{pgf}
\usepackage{setspace}
\usepackage{url}
\usepackage{hyperref}
\usepackage{bm}

\usepackage{soul,color}
\usepackage{verbatim} % for multiline comment
\usepackage[dvipsnames]{xcolor}

\usepackage{colortbl}

\usepackage{capt-of,etoolbox}
\usepackage{pifont} % Required for \ding (used in \xmark)
\usepackage{xspace}

\makeatletter
\let\NAT@parse\undefined
\makeatother
\usepackage[numbers, sort&compress]{natbib}
\usepackage{Packages/sparo_acronyms}
\usepackage{Packages/sparo_math}
\usepackage{Packages/sparo_SIunits}
\usepackage{Packages/sparo_misc}
\usepackage{multirow}
\usepackage{indentfirst}

\usepackage{soul,color}
\usepackage{verbatim} % for multiline comment
\usepackage{threeparttable} % 프리앰블에 추가

\definecolor{green}{RGB}{6,176,81}        
\definecolor{purple}{RGB}{186,117,218}    
\definecolor{red}{RGB}{204,0,0}         

\definecolor{darkyellow}{RGB}{246, 202,79}
\definecolor{best}{RGB}{198,233,201} 
\definecolor{second}{RGB}{255,247,188} 
\definecolor{trivial}{RGB}{255,200,200}

\usepackage{amssymb}
\usepackage{float}

\definecolor{gl}{HTML}{008000}

\title{\LARGE \bf
    GSAT: Geometric Traversability Estimation using Self-supervised Learning with Anomaly Detection for Diverse Terrains}

\author{Dongjin Cho$^{1}$, Miryeong Park$^{1}$, Juhui Lee$^{1}$, Geonmo Yang$^{1}$, and Younggun Cho$^{1\dagger}$% <-this % stops a space
	\thanks{This work was supported by National Research Foundation of Korea (NRF) grant (RS-2025-02217000) funded by the Korea government (MSIT) \& This work was supported by National Research Foundation of Korea (NRF) grant (RS-2025-24803365) funded by the Korea government (MSIT) \& This work was supported by Institute of Information \& communications Technology Planning \& Evaluation (IITP) grant funded by the Korea government(MSIT) (No.2022-0-00448/RS-2022-II220448) and Inha University.}
	\thanks{$^{1}$Dongjin Cho, $^{1}$Miryeong Park, $^{1}$Juhui Lee, $^{1}$Geonmo Yang and $^{1\dagger}$Younggun Cho  are with the Electrical and Computer Engineering and INHA Future Mobility IPCC, Inha University, Incheon, South Korea. {\tt\small [e-mail: d22g66, bark9757, dlwngml6635, ygm7422]@inha.edu, yg.cho@inha.ac.kr} \hfill \break
  }% 
}

\begin{document}

\maketitle
\begin{abstract}
Safe autonomous navigation requires reliable estimation of environmental traversability. Traditional methods have relied on semantic or geometry-based approaches with human-defined thresholds, but these methods often yield unreliable predictions due to the inherent subjectivity of human supervision. While self-supervised approaches enable robots to learn from their own experience, they still face a fundamental challenge: the positive-only learning problem. To address these limitations, recent studies have employed Positive-Unlabeled (PU) learning, where the core challenge is identifying positive samples without explicit negative supervision. In this work, we propose GSAT, which addresses these limitations by constructing a positive hypersphere in latent space to classify traversable regions through anomaly detection without requiring additional prototypes (e.g., unlabeled or negative). Furthermore, our approach employs joint learning of anomaly classification and traversability prediction to more efficiently utilize robot experience. We comprehensively evaluate the proposed framework through ablation studies, validation on heterogeneous real-world robotic platforms, and autonomous navigation demonstrations in simulation environments. Our method is available at \textnormal{\url{https://sparolab.github.io/research/gsat/}}.
\end{abstract}
\section{Introduction}
Traversability estimation is an essential ability to determine whether a robot can safely traverse a given terrain for autonomous navigation in unstructured environments.

Traditionally, this task has followed two main paradigms: semantic methods~\cite{hosseinpoor2021traversability, guan2022ga} employ predefined classes (e.g., road, rock) to determine traversability categories (e.g., safe, risky), whereas geometric methods~\cite{shan2018bayesian, leininger2024gaussian} extract terrain features (e.g., slope, roughness) from elevation maps. However, both methods depend on human-defined thresholds, which often yield inaccurate predictions, leading to unreliable navigation.

To address these limitations, many studies~\cite{gasparino2024wayfaster, aegidius2025watch, mattamala2025wild, cho2024learning, xue2023contrastive, seo2023learning, seo2023scate,bu2025self,jung2024v, kim2024learning} utilize self-supervised learning, enabling robots to acquire traversability knowledge from their own experience without human supervision.
In these approaches, supervision signals are automatically generated from traversal experiences. Such signals allow robots to learn models that capture platform-specific traversal patterns, thereby reducing dependence on manually designed heuristics.

\begin{figure}[t!]
    \centering
    \def\width{0.5\textwidth}%
        {%
     \includegraphics[clip, trim= 0 0 0 0, width=0.48\textwidth]{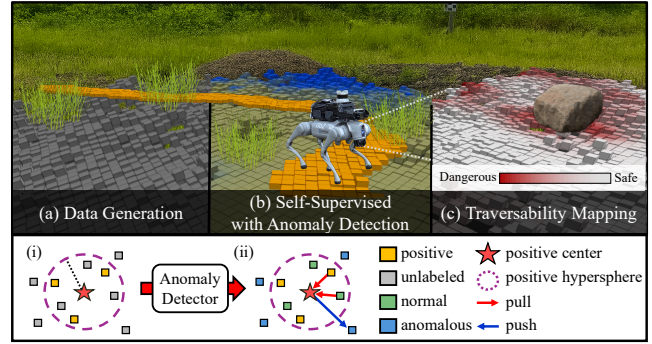}
        }
    \vspace{-0.4cm}
    \caption{\textbf{Overall process of the proposed GSAT framework:} (a) Automated data generation using positive and unlabeled samples to enable self-supervised learning, corresponding to the initial latent space (i); (b) Self-supervised anomaly detection and hypersphere refinement to optimize the decision boundary within the latent space (ii); (c) Final traversability mapping for autonomous navigation.}
    
    \vspace{-0.55cm}
    \label{fig:main}
\end{figure}

Despite this progress, self-supervised methods face the positive-only learning problem. The absence of contrastive samples (i.e., negative samples) causes the positive hypersphere representation to become unstable, leading to feature collapse, such as a trivial solution. Therefore, robots struggle to distinguish between normal samples (similar to experienced areas) and anomalous samples (unexperienced areas).

Existing methods~\cite{seo2023scate, seo2023learning, bu2025self} address this problem by constructing prototypes from positive and unlabeled sets via \ac{PU} learning, but the unlabeled data inherently contain normal samples. Consequently, the positive prototype becomes sensitive to prototype configurations (e.g., their number), resulting in inconsistent classification.

Recently, other methods~\cite{jung2024v, kim2024learning} utilize visual foundation models such as \ac{SAM}~\cite{kirillov2023segment} to mask regions similar to positive samples, treating the remaining areas as explicit negative samples. However, unlike semantic data, geometric data lacks corresponding foundation models, limiting the applicability of such foundation-based approaches.

To address this problem, we present \textbf{GSAT} (\textbf{G}eometric traversability estimation using \textbf{S}elf-supervised learning with \textbf{A}nomaly detection for diverse \textbf{T}errain), as illustrated in \figref{fig:main}. The proposed framework introduces four key contributions:
\begin{enumerate}

\item \textbf{Experience-aware Anomaly Detection}: We propose a self-supervised anomaly detection framework using a single positive hypersphere. Unlike prototype-based \ac{PU} methods, our approach defines a positive hypersphere as a decision boundary, enabling the identification of anomalies in unlabeled data. This formulation facilitates anomaly-aware metric learning, addressing the inherent limitations of positive-only learning.

\item \textbf{Joint Traversability Learning Framework}: We introduce a framework that jointly learns anomaly detection and traversability prediction. This joint optimization enables efficient terrain assessment through shared representations.

\item \textbf{Geometric Data Augmentation for Diversity}: Human-operated data often lacks diversity due to its consistent and safe operations. To address this, we introduce geometric augmentations. These generate diverse experiences, allowing the robot to adapt to various unseen regions.

\item \textbf{Comprehensive Evaluation}: 
We evaluate our model across multiple datasets for anomaly classification. 
Extensive experiments on downstream tasks demonstrate the effectiveness of our approach.

\end{enumerate}

% ===============
 
\section{Related Works} \label{sec:related_works}

\subsection{Traversability Estimation with Human Supervision}
Traditional traversability estimation relies on either semantic or geometric features. Semantic approaches~\cite{hosseinpoor2021traversability, guan2022ga} classify terrain into predefined categories (e.g., road, bush, rock) to determine traversability levels (e.g., safe, risky). Geometric methods~\cite{shan2018bayesian, leininger2024gaussian} extract terrain characteristics (e.g., slope, roughness, step height) from elevation maps generated through various probabilistic frameworks. However, these approaches depend on manually defined classes or constraints, often leading to inaccurate predictions as they fail to capture vehicle-specific traversal experiences.

\subsection{Self-Supervised Traversability Estimation}
To mitigate the reliance on human supervision, recent self-supervised approaches project driving trajectories into network features to generate automated supervision signals. Various sensor modalities have been explored for this task, including RGB~\cite{aegidius2025watch, mattamala2025wild}, RGB-D~\cite{gasparino2024wayfaster}, and LiDAR~\cite{ruetz2024foresttrav, cho2024learning}, with some methods combining modalities to leverage both geometric and semantic information~\cite{frey2024roadrunner}.

However, these methods rely primarily on reconstruction-based uncertainty derived solely from positive data. Due to the absence of contrastive samples during training, their generalization performance becomes highly sensitive to threshold configurations, often failing to establish a robust decision boundary between normal and anomalous regions.

\subsection{Traversability from Anomalies}
To address the lack of negative samples, several works adopt the \ac{PU} learning framework, treating the unlabeled set as contrastive samples~\cite{seo2023scate, seo2023learning, bu2025self}. These approaches often rely on unlabeled prototypes to distinguish between regions. However, since unlabeled sets inherently contain positive-like samples, the decision boundary becomes highly sensitive to prototype configurations, leading to inconsistent classification.

Alternative approaches~\cite{jung2024v, kim2024learning} incorporate visual foundation models, such as \ac{SAM}, to explicitly generate negative samples for refining positive prototypes. Although effective for semantic scene understanding, these methods face significant challenges in geometric domains due to the absence of corresponding foundation models.

In contrast, our method focuses solely on constructing a reliable positive prototype to define a decision boundary. This formulation enables the separation of normal and anomalous samples within unlabeled data without requiring additional prototypes or relying on foundation models. 

\begin{figure*}[t!]
% \vspace{-0.50cm}
    \centering
    \def\width{\textwidth}
    {
        \includegraphics[width=\textwidth]{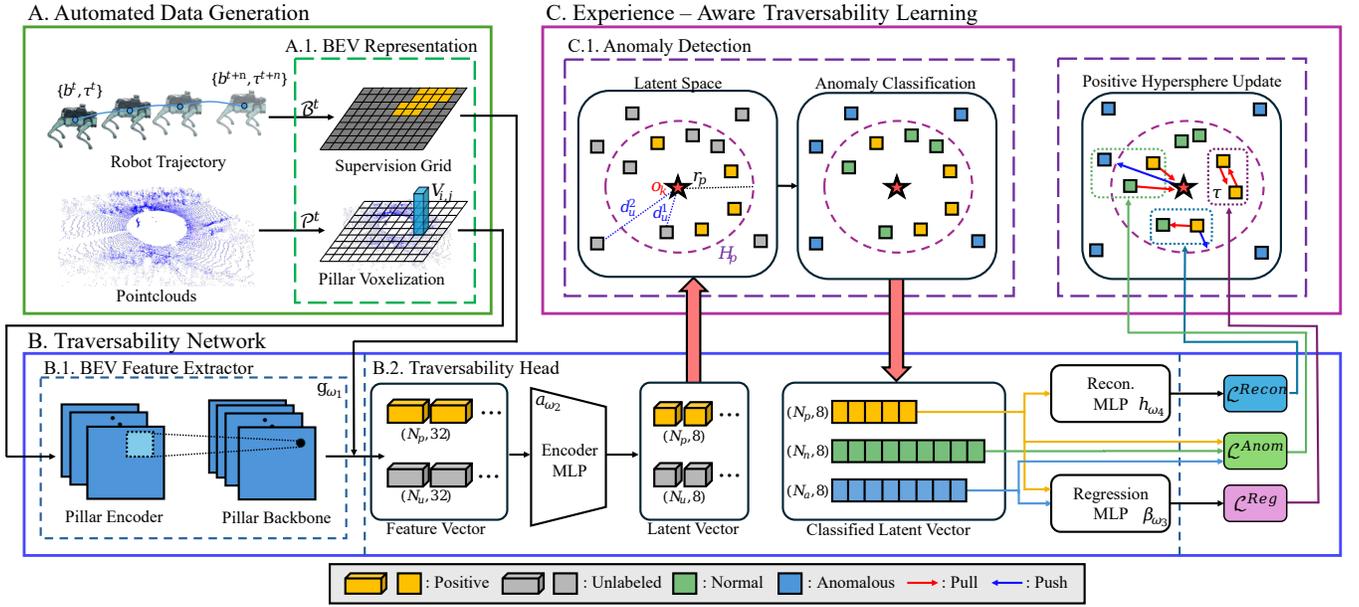}
    }
    
    \vspace{-0.5cm}
    \caption{\textbf{Overview of the proposed GSAT framework.} The framework consists of (A) automated data generation leveraging robot traversal supervision, and (B) a traversability network that extracts \ac{BEV} features to produce latent representations. The core innovation lies in (C) experience-aware traversability learning, where self-supervised anomaly detection in the latent space partitions unlabeled data into normal and anomalous samples, enabling joint traversability estimation and positive hypersphere refinement through the proposed loss formulations.}
\vspace{-0.5cm}
    \label{fig:frame_work}
\end{figure*}

\section{Method} \label{Method}
In this section, we describe how geometric information can be utilized for traversability estimation through self-supervised learning. First, we introduce an automated data generation pipeline (Sec.~\ref{auto_data_gen}). We then describe the traversability network architecture (Sec.~\ref{sec:sec_2}), followed by our proposed experience-aware anomaly detection methodology and corresponding loss functions for constructing a positive hypersphere through joint learning (Sec.~\ref{sec::sec_c}). These core components are illustrated in~\figref{fig:frame_work}. Finally, we present a targeted geometric augmentation strategy to address the limited diversity of positive samples (Sec.~\ref{auto_data_gen:pointaug}).

% =================== 1. Automated Data Generation =================== %
\subsection{Automated Data Generation} \label{auto_data_gen}
As the initial step, training data is automatically generated by aligning SLAM-derived robot trajectories with the point cloud $\mathcal{P}^t$ captured at discrete time intervals. Specifically, the system constructs this supervision by associating sampled trajectory points with traversability scores. For each time step $t$, the supervision set $\mathcal{B}^t$ is defined as:
\begin{equation}
    \mathcal{B}^t = \{ (\mathbf{b}^s, \tau^s) \mid s \in [t, t+n] \},
\end{equation}
where $\mathbf{b}^s = (x^s, y^s)$ represents the robot's position coordinates, $\tau^s$ denotes the traversability score at sampling step $s$, and $n$ denotes the window size. Traversability scores are derived from velocity tracking performance as follows~\cite{mattamala2025wild}:
\begin{equation}
    v_{\text{error}} = \frac{1}{2}\left((v_x - v_{x^*})^2 + (v_y - v_{y^*})^2\right),
\end{equation}
\begin{equation}
    \tau = \sigma\left(-\eta(v_{\text{error}} - v_{\text{th}})\right),
\end{equation}
\noindent where $(v_x, v_y)$ and $(v_{x^*}, v_{y^*})$ are the actual and commanded velocities, respectively. The parameter $\eta$ controls the sharpness of the sigmoid function $\sigma(\cdot)$, and $v_{\text{th}}$ denotes the midpoint where the traversability score is assigned as $0.5$.

\subsubsection{BEV Representation} \label{auto_data_gen:BEV_representation}
Direct processing of 3D point clouds limits performance on resource-constrained robotic platforms~\cite{lang2019pointpillars}. Therefore, we adopt a \ac{BEV} representation to achieve computational efficiency for real-time applications. To align $\mathcal{P}^t$ and $\mathcal{B}^t$, both are projected onto their respective 2D grids of size $H_\rho \times W_\rho$ with resolution $\rho$.

\noindent\textbf{Pillar Voxelization:}
$\mathcal{P}^t$ is voxelized into a 4D tensor $\mathcal{V} \in \mathbb{R}^{H_\rho \times W_\rho \times M \times 7}$, where each grid cell $(i,j)$ stores a vertical pillar $\mathbf{V}_{i,j} = \{\mathbf{v}_{i,j}^{(m)}\}_{m=1}^M \in \mathbb{R}^{M \times 7}$. $M$ denotes the maximum number of points per pillar, with the remaining entries zero-padded. Each per-point feature $\mathbf{v}_{i,j}^{(m)} \in \mathbb{R}^7$ is defined as:
\vspace{-0.4em}
\begin{equation}
\mathbf{v}_{i,j}^{(m)} = [x_l, y_l, z, x_c, y_c, z_c, \sigma_z]^\top,
\end{equation}
where $z$ is the absolute height of the point, $(x_l, y_l)$ denote the local offsets from the pillar center, $(x_c, y_c, z_c)$ are the offsets relative to the mean of all points in the pillar, and $\sigma_z$ is the standard deviation of $z$ within the pillar.

\noindent\textbf{Supervision Grid:}
The supervision set $\mathcal{B}^t$ is represented as a 2D grid $\mathbf{G} \in \mathbb{R}^{H_\rho \times W_\rho}$. Each cell $\mathbf{G}_{i,j}$ containing a robot position $\mathbf{b}^s$ stores the corresponding traversability score $\tau^s$. When multiple positions occupy the same cell, the mean score is assigned. Unvisited cells are set to $-1$.

% =================== 2. Traversability Network =================== %
\subsection{Traversability Network} \label{sec:sec_2}
\subsubsection{BEV Feature Extractor}
We design our feature extractor based on the PointPillars~\cite{lang2019pointpillars} architecture. Pillar features $\mathbf{V}_{i,j}$ are encoded through a 1D-CNN followed by a 2D-CNN backbone, producing a spatial feature vector $\mathbf{q}_{i,j} \in \mathbb{R}^{32}$ for each grid cell $(i,j)$. The BEV feature map $\mathbf{Q}$ is composed of per-cell feature vectors $\mathbf{q}_{i,j}$, which are computed as:
\begin{equation}
    \mathbf{q}_{i,j} = g_{\omega_1}(\mathbf{V}_{i,j}),
\end{equation}
where $g_{\omega_1}(\cdot)$ denotes the BEV feature extraction module with learnable parameters $\omega_1$. The feature vectors are separated into positive and unlabeled sets based on the supervision grid $\mathbf{G}$:
\begin{equation}
\begin{split}
    \mathcal{Q}_p &= \{\mathbf{q}_{i,j} \mid \mathbf{G}_{i,j} \neq -1\}, \\
    \mathcal{Q}_u &= \{\mathbf{q}_{i,j} \mid \mathbf{G}_{i,j} = -1\},
\end{split}
\end{equation}
\noindent where the subscripts $p$ and $u$ indicate the positive and unlabeled sets, respectively.

\subsubsection{Traversability Head}
The Traversability Head processes the BEV feature set $\mathcal{Q} = \mathcal{Q}_p \cup \mathcal{Q}_u$ through three MLPs:
\begin{equation}
\mathcal{Z} = \alpha_{\omega_2}(\mathcal{Q}), \quad
\mathcal{T} = \beta_{\omega_3}(\mathcal{Z}), \quad
\mathcal{U} = h_{\omega_4}(\mathcal{Z}),
\end{equation}
where $\alpha_{\omega_2}(\cdot)$ denotes the encoder, $\beta_{\omega_3}(\cdot)$ the regression head, and $h_{\omega_4}(\cdot)$ the reconstruction head. Specifically, for $N = N_p + N_u$ total samples where $N_p = |\mathcal{Q}_p|$ and $N_u = |\mathcal{Q}_u|$, $\mathcal{Z} \in \mathbb{R}^{N \times 8}$ denotes the latent representations, $\mathcal{T} \in \mathbb{R}^{N \times 1}$ the predicted traversability scores, and $\mathcal{U} \in \mathbb{R}^{N \times 32}$ the reconstructed features. The encoder and reconstruction heads consist of two layers each, while the regression head has a single layer. All hidden layers employ batch normalization, dropout, and ReLU, whereas the final layer of the regression head applies a sigmoid activation.

% =================== 3. Anomaly Detection && Loss =================== %
\subsection{Experience-Aware Traversability Learning}\label{sec::sec_c}
\subsubsection{Anomaly Detection}\label{sec::sec_c::anomaly}
Unlike methods~\cite{seo2023scate, seo2023learning, bu2025self, jung2024v, kim2024learning} that rely on unlabeled or negative prototypes, we employ a metric-based approach using positive hypersphere boundaries to separate normal and anomalous samples in unlabeled data.

The positive hypersphere $H_p$ is defined by its center $\mathbf{o}_k \in \mathbb{R}^8$ and radius $r_p$. The center $\mathbf{o}_k$ is defined as the mean feature vector of positive latent features $\mathbf{z}_p^{(k)}$, computed over the full batch and updated every $k$ epochs for stable convergence:
\begin{equation}
\mathbf{o}_k = \frac{1}{N_p} \sum_{\mathcal{Z}_p} \mathbf{z}_p^{(k)}.
\end{equation}
The radius $r_p$ follows an exponential moving average:
\begin{equation}
r_p^+ = \varepsilon r_p^- + (1-\varepsilon) \bar{d}_p,
\end{equation}
\noindent where $d_p = \lVert \mathbf{z}_p - \mathbf{o}_k \rVert$ is the distance between each positive latent vector $\mathbf{z}_p \in \mathcal{Z}_p$ and the center, $\bar{d}_p$ is their mean, and $\varepsilon$ is the momentum parameter. Superscripts $-$ and $+$ denote pre- and post-update values at each epoch. Using $H_p$ as a boundary, unlabeled samples $\mathbf{z}_u \in \mathcal{Z}_u$ are split into normal $\mathcal{Z}_n$ and anomalous $\mathcal{Z}_a$ sets:
\vspace{-0.1em}
\begin{equation}
\begin{split}
\mathcal{Z}_n &= \left\{ \mathbf{z}_u \in \mathcal{Z}_u \mid d_u \leq r_p^+ \right\}, \\
\mathcal{Z}_a &= \left\{ \mathbf{z}_u \in \mathcal{Z}_u \mid d_u > r_p^+ \right\},
\end{split}
\label{equation:classification}
\end{equation}
where $d_u = \lVert \mathbf{z}_u - \mathbf{o}_k \rVert$ represents the distance from each unlabeled sample to the hypersphere center, with $N_n = |\mathcal{Z}_n|$ and $N_a = |\mathcal{Z}_a|$ denoting the sizes of each set, satisfying \mbox{$N_u = N_n + N_a$}.

\subsubsection{End-to-End Training Objective} \label{sec:sec_3}
To address the positive-only problem in traversability estimation, we jointly optimize traversability prediction and anomaly-aware representation learning. Specifically, the overall training objective consists of the loss functions detailed below.

\noindent \textbf{Anomaly Loss.}
We focus on optimizing a positive prototype without negative or unlabeled ones. Therefore, we adapt an anomaly loss inspired by Deep-SAD~\cite{ruff2019deep}, which refines the positive center by explicitly pushing anomalous samples away from it. While Deep-SAD requires pre-labeled normal and anomalous data, our approach leverages automated anomaly detection (Sec.~\ref{sec::sec_c::anomaly}) to classify unlabeled samples, enabling the application of this loss formulation. The anomaly loss is defined as follows:

\vspace{-1.3em}
\begin{equation}
\begin{aligned}
    \mathcal{L}_{\text{Anom}} =\;&
    \underbrace{\frac{1}{N_p} \sum_{\mathcal{Z}_p} \|\mathbf{z}_p - \mathbf{o}_k\|^2}_{\text{positive}}
    + \underbrace{\frac{1}{N_n} \sum_{\mathcal{Z}_n} \|\mathbf{z}_n - \mathbf{o}_k\|^2}_{\text{normal}} \\
    &+ \underbrace{\frac{1}{N_a} \sum_{\mathcal{Z}_a} \left(\|\mathbf{z}_a - \mathbf{o}_k\|^2 + \zeta\right)^{-1}}_{\text{anomalous}},
\end{aligned}
\end{equation}
\vspace{-0.5em}

\noindent where $\mathbf{z}_p \in \mathcal{Z}_p$, $\mathbf{z}_n \in \mathcal{Z}_n$, and $\mathbf{z}_a \in \mathcal{Z}_a$ are latent vectors from the positive, normal, and anomalous sets, respectively. The first two terms pull positive and normal samples toward the hypersphere center, while the third term pushes anomalous samples away through inverse distance. $\zeta$ is a small constant that prevents numerical divergence. This formulation constructs a positive boundary that enables anomaly detection without requiring auxiliary prototypes.

\noindent \textbf{Reconstruction Loss.} 
While the anomaly loss effectively learns hypersphere boundaries, pulling positive features toward the center without any margin may cause the encoder to overfit to specific positive patterns. To mitigate this, we employ a reconstruction loss formulated as the \ac{MSE} to preserve general representations:
\begin{equation}
    \mathcal{L}_{\text{Recon}} = \frac{1}{N_p} \sum_{\mathcal{U}_p} \|\mathbf{u}_p - \mathbf{q}_p\|^2,
\end{equation}
\noindent where $\mathbf{u}_p \in \mathcal{U}_p$ represents the reconstructed feature decoded from the latent vector $\mathbf{z}_p$, and $\mathbf{q}_p$ denotes the corresponding original positive input. As a regularizer, this loss complements the anomaly loss by preventing feature collapse and ensuring robust representations for anomaly detection.

\noindent \textbf{Regression Loss.} 
Our regression loss leverages experience-aware joint learning to achieve efficient traversability estimation. Existing methods~\cite{mattamala2025wild} rely on reconstruction error as a heuristic anomaly indicator with manually defined thresholds. Unlike these methods, our approach utilizes the classified samples, eliminating the need for such thresholds. We formulate the regression objective using the \ac{MSE} as follows:
\begin{equation}
    \mathcal{L}_{\text{Reg}} = \frac{1}{N_p} \sum_{\mathcal{T}_p} \|\mathbf{t}_p - \tau\|^2 + \frac{1}{N_a} \sum_{\mathcal{T}_a} \|\mathbf{t}_a - 0\|^2,
\end{equation}
\noindent where $\mathbf{t}_p \in \mathcal{T}_p$ and $\mathbf{t}_a \in \mathcal{T}_a$ denote the traversability predictions for positive and anomalous samples, respectively. The first term supervises $\mathbf{t}_p$ toward the target score $\tau$, while the second term constrains $\mathbf{t}_a$ toward zero to reflect risk regions. This joint formulation enables efficient traversability estimation.

\noindent \textbf{Total Loss.}
The final objective function combines all three loss components as follows:
\begin{equation}
    \mathcal{L}_{\text{Total}} = \lambda_1 \mathcal{L}_{\text{Anom}} + \lambda_2 \mathcal{L}_{\text{Recon}} + \lambda_3 \mathcal{L}_{\text{Reg}},
\end{equation}
\noindent where $\lambda_1$, $\lambda_2$, and $\lambda_3$ weight the respective loss terms.

% ================= Data Augmentation =================
\begin{figure}[t!]
    \centering
    \def\width{\textwidth}%
        {
        \includegraphics[clip, trim= 0 0 0 0, width=0.48\textwidth]{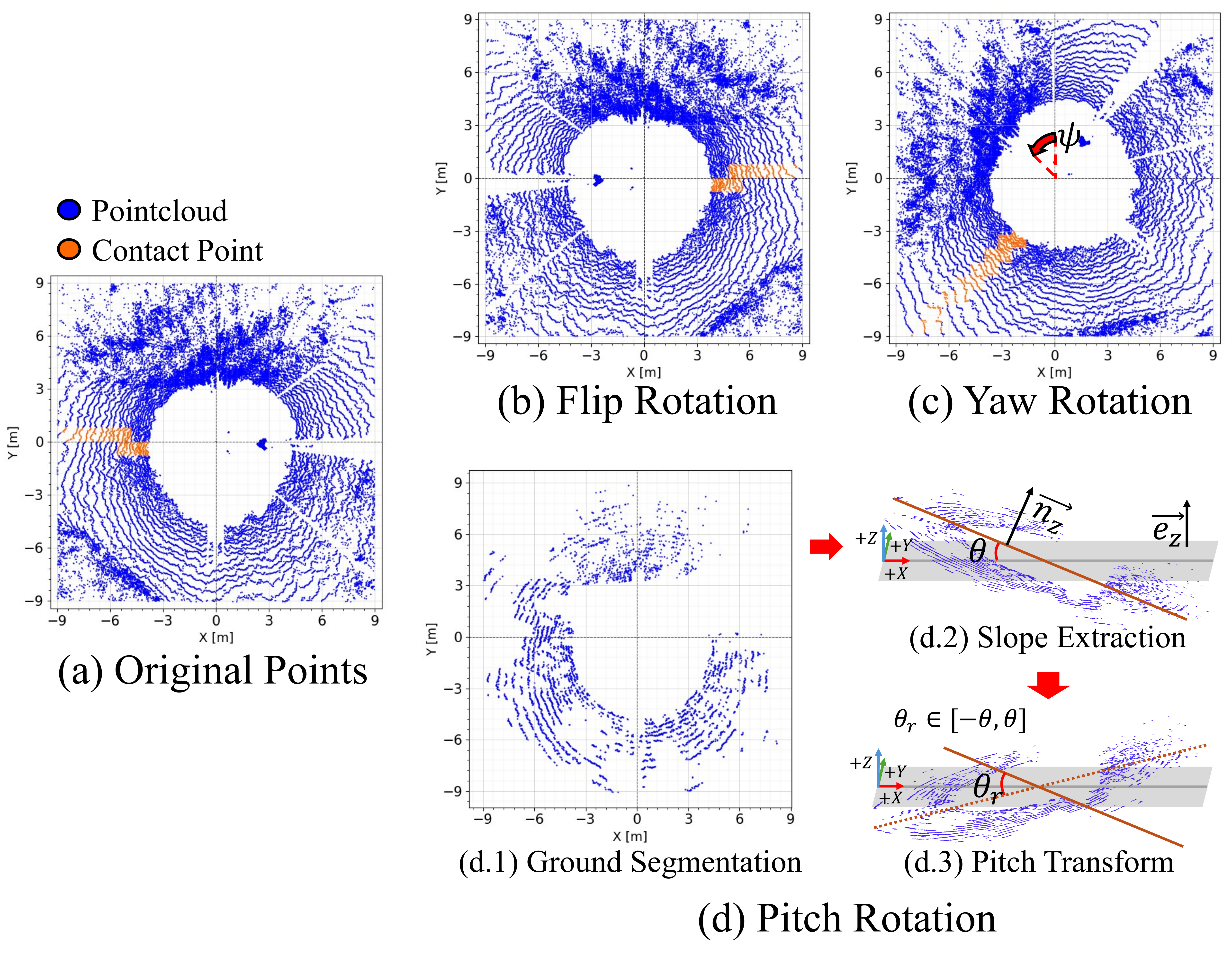}
        }
    \vspace{-0.5cm}
    \caption{Point augmentation methods demonstrated on RELLIS-3D dataset. (a) Sample point clouds. (b) Flipping augmentation reflecting points across the $yz$-plane. (c) Yaw rotation augmentation around the $z$-axis. (d) Pitch augmentation process: (d.1) RANSAC-based ground segmentation, (d.2) slope angle extraction, (d.3) pitch transformation to simulate terrain inclinations.}
    \vspace{-0.3cm}
    \label{fig:aug_method}
\end{figure}
%================================================
\subsection{Strategy for Mitigating Limited Data Diversity} \label{auto_data_gen:pointaug}
While our anomaly detection framework expands the positive hypersphere using normal samples from unlabeled data, the positive samples themselves often exhibit limited diversity due to consistent human operational patterns. Therefore, we apply targeted geometric augmentations to mitigate orientation and terrain biases inherent in positive traversal data. We employ flipping and yaw rotation to address directional bias, and pitch rotation based on ground estimation to increase slope diversity. The detailed augmentation strategies are as follows:

\noindent \textbf{Flipping.} As shown in \figref{fig:aug_method}(b), we flip points across the $yz$-plane using a diagonal matrix $\mathbf{S}_{x} = \text{diag}(-1, 1, 1)$. The transformation $f_{x}(\mathbf{p}) = \mathbf{S}_{x}\mathbf{p}$ is applied to each point $\mathbf{p} = [x, y, z]^\top$, creating symmetric trajectory patterns and mitigating distributional bias in unidirectional sampling.

\noindent \textbf{Yaw Rotation.} As shown in \figref{fig:aug_method}(c), we apply random yaw rotations about the $z$-axis using a rotation matrix $\mathbf{R}_{z}(\psi)$, where $\psi \in [-90^\circ, 90^\circ]$. This transformation $f_\psi(\mathbf{p}) = \mathbf{R}_{z}(\psi)\mathbf{p}$ enhances angular diversity.

\noindent \textbf{Pitch Rotation.} As shown in \figref{fig:aug_method}(d), we simulate terrain slope variations through pitch rotation based on RANSAC ground segmentation. The ground slope angle $\theta = \arccos(\mathbf{n}_z \cdot \mathbf{e}_z)$ determines the rotation range, where $\mathbf{n}_z$ is the estimated ground normal and $\mathbf{e}_z = [0, 0, 1]^\top$ is the vertical unit vector. Assuming that any slope up to $\theta$ remains traversable, a random angle $\theta_r \in [-\theta, \theta]$ is then applied as $f_{g}(\theta_r, \mathbf{p}) = \mathbf{R}_y(\theta_r)\mathbf{p}$, generating diverse slope conditions.

\section{Experiments on Anomaly Classification} \label{Experiments on Anomaly Classification}
We conducted comprehensive ablation studies to analyze the impact of two design choices within our anomaly classification: the treatment of unlabeled data in the anomaly loss and the selection of augmentation strategies. Following the experimental setup, including datasets and evaluation metrics (Sec.~\ref{experi_1:setup}), we quantitatively analyze the anomaly loss formulation (Sec.~\ref{experi_1:anomaly_ab}) and evaluate each augmentation component through both quantitative and qualitative assessments (Sec.~\ref{experi_1:augment_ab}).

\subsection{Experimental Setup} \label{experi_1:setup}

\subsubsection{Datasets}
We conducted experiments on two public datasets: RELLIS-3D~\cite{jiang2021rellis}, which features consistent positive sample patterns, and DITER++~\cite{kim2024diter++}, which provides diverse spatial environments. For both datasets, we sampled approximately 50 robot poses per training to define the positive samples.

\noindent\textbf{RELLIS-3D}: This dataset was collected with an OS1-64 LiDAR sensor in off-road environments and provides point-wise semantic annotations. We used sequences 00 and 03 for training and sequence 01 for evaluation. For anomaly evaluation, we redefined the provided semantic labels into traversability categories. Specifically, common terrain types (e.g., \textit{ground, grass, mud, and puddles}) were grouped as normal, whereas all other labels (e.g., \textit{trees, rocks}) were treated as anomalous.

\noindent\textbf{DITER++}: This dataset was collected with an OS1-32 LiDAR sensor in diverse urban and park environments. Training was conducted on the park-night sequence and evaluation on the park-day sequence. For evaluation, since semantic annotations were not provided, we manually labeled non-ground and obstacle regions as anomalous, while areas consistent with prior experience were labeled as normal.

\subsubsection{Data Processing and Evaluation Metrics}
We utilized LiDAR point clouds, extracting supervision signals from SLAM-based robot poses. Since command velocity information was unavailable, we set the target traversability score to $\tau=1$. To evaluate anomaly classification, point-wise semantic labels were projected onto a \ac{BEV} grid of $12 \times 12$ m with a 0.15 m resolution. Cells containing any points from predefined anomalous classes were labeled as anomalous, while cells with only normal class points were labeled as normal. 

Anomaly classification was performed using a decision boundary determined by the learned hypersphere center and radius, as defined in Eq.~\eqref{equation:classification}. We evaluated binary classification performance using Precision, Recall, and F1-score. To focus on regions with reliable measurements, empty cells without sufficient point cloud data were excluded from the evaluation. In \tabref{tab:anomaly_loss_ablation} and \tabref{tab:augmentation_results}, green and yellow highlights denote the best and second-best results, respectively.

\subsubsection{Implementation Details}
The network was trained using the Adam optimizer with a learning rate of $5 \times 10^{-4}$, a batch size of 12, and 100 training epochs. The hypersphere parameters were updated every 5 epochs ($k=5$) using a momentum of $\varepsilon = 0.5$. For numerical stability, we set $\zeta = 10^{-6}$ in the anomaly loss term. For the classification evaluation, the hypersphere center and radius were selected from the best-performing model on the evaluation set. The loss weights were empirically set to $\lambda_1=1.0$, $\lambda_2=1.0$, and $\lambda_3=20.0$. All models were trained and evaluated on an NVIDIA RTX 3090 GPU with 24GB of memory. To ensure reliability, the results were averaged over 3 independent runs with different random seeds.

\subsection{Anomaly Loss Ablation Studies} \label{experi_1:anomaly_ab}
\tabref{tab:anomaly_loss_ablation} presents a quantitative evaluation of four configurations on the RELLIS-3D and DITER++ datasets. Specifically, we analyze how the handling of unlabeled data is formulated within the anomaly loss.

\begin{figure*}[t!]
% \vspace{-0.50cm}
    \centering
    \def\width{\textwidth}
    {
        \includegraphics[width=\textwidth, clip, trim=15mm 15mm 15mm 15mm]{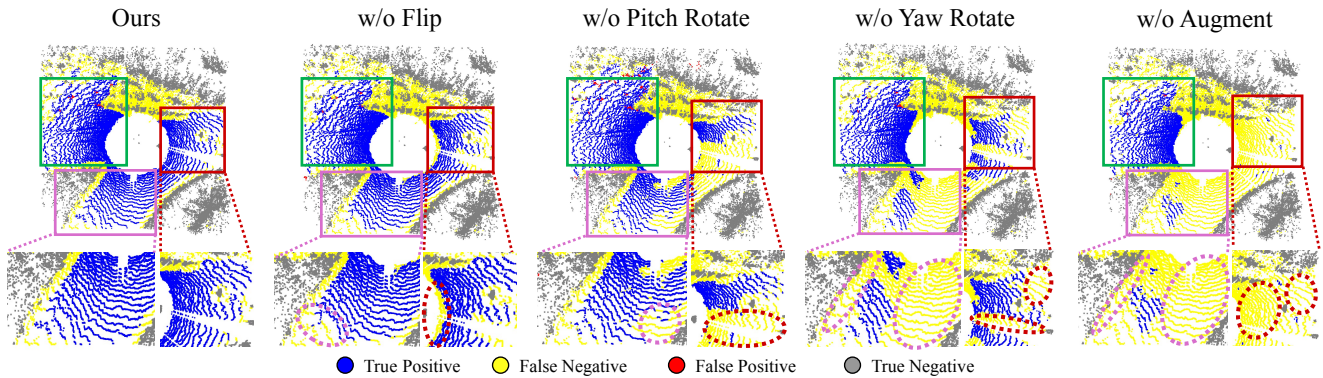}
    }
    \vspace{-0.4cm}
    \caption{Qualitative results of the data augmentation ablation study on the RELLIS-3D dataset, demonstrating spatial prediction performance across different augmentation configurations.}
    \vspace{-0.10cm}
    \label{fig:aug_qualitative}
\end{figure*}

\begin{table}[t!]
\centering
\caption{Quantitative results of the unlabeled data ablation study for anomaly loss formulation} 
\label{tab:anomaly_loss_ablation}
\resizebox{0.48\textwidth}{!}{%
\begin{tabular}{c|cc|ccc}
\toprule
\toprule
\multirow{2}{*}{\textbf{Dataset}} 
 & \multicolumn{2}{c|}{\hspace{0.8em}\textbf{Anomaly Loss}\hspace{0.8em}} 
 & \multicolumn{3}{c}{\hspace{0.3em}\textbf{Metric}\hspace{0.3em}} \\
\cline{2-3} \cline{4-6}
 & Normal & Anomalous & Pre.\,$\uparrow$ (\%) & Rec.\,$\uparrow$ (\%) & F1.\,$\uparrow$ (\%) \\
\midrule
\midrule
\multirow{4}{*}{RELLIS-3D}
 & --       & --              & 54.75 & \cellcolor{best}99.86 & 69.87 \\
 & --       & All unlabeled   & 96.75 & 52.52 & 68.27 \\
 & --       & Anomalous  & \cellcolor{best}98.10 &59.14 & \cellcolor{second}73.59 \\
 & Normal   & Anomalous       & \cellcolor{second}97.11 & \cellcolor{second}64.93 & \cellcolor{best}77.61 \\
\midrule
\multirow{4}{*}{DITER++}
 & --       & --              & 77.34 & \cellcolor{best}98.35 & \cellcolor{second}85.91 \\
 & --       & All unlabeled   & 87.45 & 57.56 & 68.50 \\
 & --       & Anomalous  & \cellcolor{second}87.69 & 75.93 & 80.61 \\
 & Normal   & Anomalous       & \cellcolor{best}96.69 & \cellcolor{second}81.01 & \cellcolor{best}88.04 \\
\bottomrule
\bottomrule
% \multicolumn{6}{l}{\fontsize{10pt}{10pt}\selectfont Ours: Normal/Anomalous configuration}\\[-0.6em]
\multicolumn{6}{l}{\fontsize{10pt}{10pt}\selectfont Ours: Normal/Anomalous configuration}
\end{tabular}}
\vspace{-0.4cm}
\end{table}

\noindent \textbf{w/o Anomaly Loss ($-$/$-$):} This configuration, equivalent to Deep-SVDD~\cite{ruff2019deep} one-class learning, predicted most regions as normal, classifying even non-traversable areas as traversable. Consequently, it yielded the highest recall (99.86\%, 98.35\%) but the lowest precision (54.75\%, 77.34\%). Without loss terms for unlabeled data, the model failed to distinguish between normal and anomalous latent features, leading to inflated recall scores that do not reflect true classification performance.

\noindent \textbf{All Unlabeled as Anomalous ($-$/All unlabeled):} This configuration treated all unlabeled data as anomalous, pushing them away from the positive hypersphere center. Both datasets exhibited significantly reduced recall (52.52\%, 57.56\%) because the model overfitted to positive patterns, creating an overly restrictive hypersphere that missed normal regions in unexplored areas. While this approach achieved high precision (96.75\%, 87.45\%) by being conservative, it failed to generalize beyond experienced terrain.

\noindent \textbf{Anomalous-only Loss ($-$/Anomalous only):} This configuration constructed the hypersphere using only positive samples while pushing anomalous samples away from the center, without utilizing normal samples. This approach demonstrated an improved balance between precision (98.10\%, 87.69\%) and recall (59.14\%, 75.93\%). However, learning the hypersphere based solely on positive samples limited its generalization capability, as normal samples were not guided toward the center.

\noindent \textbf{Proposed Configuration (Normal/Anomalous):} Our approach optimized the positive hypersphere by pulling both positive and normal samples toward the center. This incorporation of normal samples significantly improved performance compared to positive-only approaches, achieving the highest F1-scores (77.61\%, 88.04\%). These results validated our hypothesis that leveraging normal samples enhances hypersphere construction and improves anomaly detection performance.

\begin{table}[t!]
\centering
\caption{Quantitative results of data augmentation on RELLIS-3D}
\label{tab:augmentation_results}
\scalebox{0.8}{
\begin{tabular}{l|ccc}
\toprule
\toprule
\textbf{Configuration} & \textbf{Pre.\,$\uparrow$ (\%)} & \textbf{Rec.\,$\uparrow$ (\%)} & \textbf{F1.\,$\uparrow$ (\%)} \\
\midrule
\midrule
w/o Flipping          & 96.67 & \cellcolor{second}61.32 & \cellcolor{second}74.86 \\
w/o Pitch rotation  & 93.82 & 61.72 & 74.16 \\
w/o Yaw rotation    & 96.78 & 45.54 & 61.81 \\
w/o Augmentation    & \cellcolor{best}98.28 & 27.99 & 43.13 \\
\midrule
\textbf{Ours}       & \cellcolor{second}97.11 & \cellcolor{best}64.93 & \cellcolor{best}77.61 \\
\bottomrule
\bottomrule
\end{tabular}}
\vspace{-0.3cm}
\end{table}

The performance difference between the datasets reflected ground-truth annotation quality. RELLIS-3D employed semantic class-based labeling with inherent inaccuracies (e.g., grass regions were often labeled as traversable despite being non-traversable due to their irregular surfaces). Consequently, achieving high precision on this dataset is a more meaningful indicator of performance than recall. In contrast, DITER++ used more precise manual ground-only labeling, resulting in clearer ground-truth classification and consistently higher overall performance across all configurations.

\subsection{Ablation Studies on Geometric Augmentation}
\label{experi_1:augment_ab}
We conducted augmentation ablation studies on the RELLIS-3D dataset to evaluate the effectiveness of our geometric augmentation strategies. Unlike DITER++, RELLIS-3D exhibits highly consistent trajectory patterns and terrain biases (e.g., directional and slope biases), making it an ideal benchmark for validating the impact of augmentation. Furthermore, to avoid generating misleading positive samples, pitch rotation was selectively applied only to scenes containing a single ground segment with a slope of less than $10^\circ$.

\tabref{tab:augmentation_results} presents the quantitative evaluation of each augmentation component. Our augmentation approach achieved the highest F1-score (77.61\%), with each technique contributing significantly to performance. Notably, excluding yaw rotation resulted in the most severe performance drop with an F1-score of 61.81\%, demonstrating its critical importance for handling directional diversity. Without any augmentation, the model achieved high precision (98.28\%) but suffered from extremely low recall (27.99\%), indicating overfitting to the limited positive training patterns and weak generalization to unexplored regions.

\figref{fig:aug_qualitative} presents qualitative results showing the effect of each augmentation on prediction performance. Without augmentation, the model performed reliably only in regions similar to training trajectories but failed in unexplored orientations, showing extensive false negatives in the purple and red highlighted regions. The absence of yaw rotation particularly affected predictions in unseen headings, while removing pitch rotation degraded performance on sloped terrain (highlighted in red). Specifically, because the positive training data exhibited a leftward bias, the model severely struggled with rightward direction patterns without flip augmentation. These results demonstrate that geometric augmentation effectively addresses the inherent biases in training data, improving generalization to diverse terrains.

% ================= setup =================
\begin{figure}[t!]
    % \vspace{-0.5cm}
    \centering
	\def\width{0.48\textwidth}%
        {%
		\includegraphics[clip, trim= 0 0 0 0, width=\width]{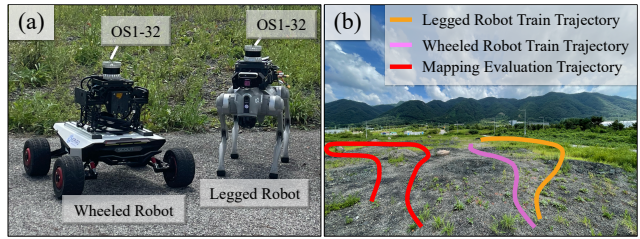}
	}
    \vspace{-0.6cm}
    \caption{Experimental setup for traversability mapping evaluation. (a) Robot platforms: SCOUT MINI (left) and Go2 (right). (b) Data collection trajectories in the outdoor environment with training and mapping evaluation paths.}
    \label{fig:custom_dataset_setup}
    \vspace{-0.5cm}
\end{figure}
%================================================
\section{Experiments on Downstream Tasks}
To evaluate our experience-aware traversability estimation, we compared it against representative open-source baselines. Specifically, two methods were selected for comparison: DEM-Trav~\cite{leggedrobotics_traversability_estimation}, a rule-based approach that assesses geometric terrain characteristics, and LeSTA~\cite{cho2024learning}, a self-supervised method that learns terrain properties. Platform-specific traversability accuracy was evaluated through mapping experiments (Sec.~\ref{experi_2:mapping}), and its suitability for navigation in unstructured environments was assessed in simulation (Sec.~\ref{experi_2:nav}).

\begin{figure}[t!]
    % \vspace{-0.5cm}
    \centering
	\def\width{0.48\textwidth}%
        {%
		\includegraphics[clip, trim= 0 0 0 0, width=\width]{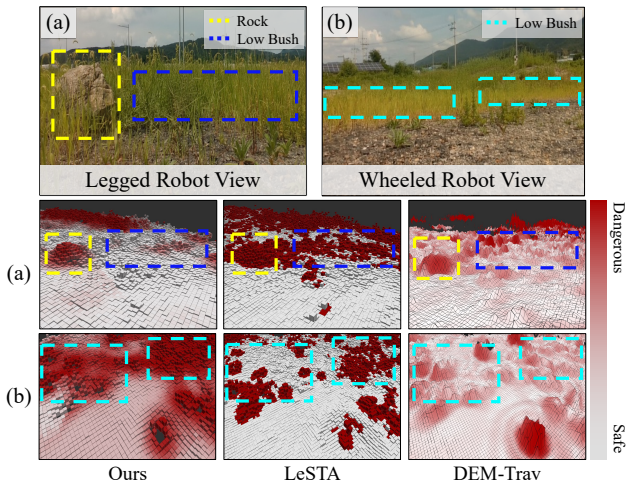}
	}
    \vspace{-0.8cm}
    \caption{Traversability mapping results comparing benchmark methods in the real-world. (a) Legged robot: low bush regions are traversable, rock regions are dangerous. (b) Wheeled robot: low bush regions are non-traversable.}
    \label{fig:travresability_mapping_bench}
    \vspace{-0.7cm}
\end{figure}

\subsection{Traversability Mapping}\label{experi_2:mapping}
\subsubsection{Experimental Setup} 
We evaluated traversability mapping across heterogeneous robots using a custom dataset collected from wheeled and legged robots equipped with OS1-32 LiDAR sensors, as shown in \figref{fig:custom_dataset_setup}(a). The environment consisted of three terrain types: uneven road, low bush ($< 0.6$ m), and high bush ($\geq 0.6$ m). Based on assumed platform-specific capabilities, the legged robot was assigned to traverse \textit{uneven road} and \textit{low bush}, while the wheeled robot was restricted to \textit{uneven road}. Each robot's trajectory was split into 80\% for training and 20\% for validation. For testing, both robots navigated an identical evaluation trajectory, as shown in \figref{fig:custom_dataset_setup}(b). 

Our approach utilized the training parameters detailed in Sec.~\ref{experi_1:setup}. For LeSTA, we used input features (step-height, slope, roughness, and curvature) with a learning rate of 0.7. For DEM-Trav, we employed step-height, roughness, and slope features with robot-specific maximum thresholds: 1.4 m, 0.2 m, and 1.0 rad for the legged robot, and 0.6 m, 0.05 m, and 1.0 rad for the wheeled robot, respectively. All methods used a grid size of $8 \times 8$ m with a 0.15 m resolution.

\subsubsection{Traversability Mapping Results}

\begin{figure}[t!]
    % \vspace{-0.5cm}
    \centering
	\def\width{0.48\textwidth}%
        {%
		\includegraphics[clip, trim= 0 0 0 0, width=\width]{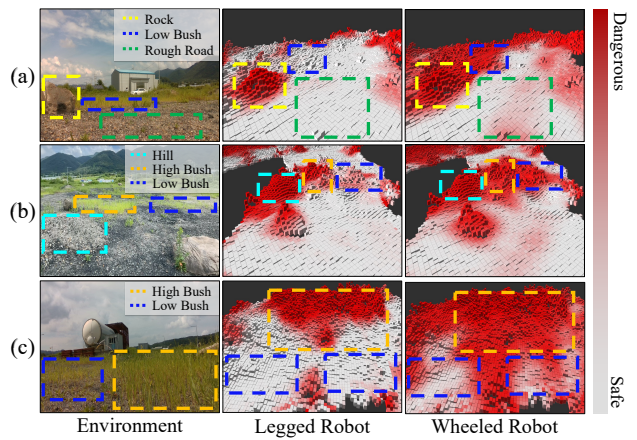}
	}
    \vspace{-0.8cm}
    \caption{Detailed traversability mapping results of the proposed method, illustrating robot-specific terrain assessment for legged and wheeled robots.}
    \label{fig:travresability_mapping}
    % \vspace{-0.5cm}
\end{figure}

As shown in \figref{fig:travresability_mapping_bench} and \figref{fig:travresability_mapping}, our method successfully generated robot-specific traversability maps. For the legged robot, low bush regions were correctly identified as traversable, whereas for the wheeled robot, bush areas were appropriately predicted as non-traversable. Both robots consistently predicted rock and hill regions as non-traversable. Notably, the regions highlighted in green (\figref{fig:travresability_mapping}) demonstrate distinct traversability scores for identical uneven road surfaces, accurately reflecting the wheeled robot's kinematic limitations by assigning higher risk to such terrain.

In contrast, both LeSTA and DEM-Trav exhibited limited adaptability across platforms. LeSTA predicted low bushes as non-traversable for the legged robot and incorrectly estimated high bushes as traversable for the wheeled robot. Similarly, DEM-Trav predicted low bushes as high-risk terrain for the legged robot and also failed to handle dense bushes for the wheeled robot. Ultimately, DEM-Trav relies on human-defined capability thresholds that may not accurately reflect platform-specific mobility constraints, whereas LeSTA, by focusing on terrain properties, may fail to fully capture robot-dependent traversability characteristics.

\begin{table}[t]
  \centering
  \caption{Navigation Performance Comparison}
  \label{tab:mission_overall}
  \scalebox{0.8}{
  \begin{tabular}{l|cc}
    \toprule
    \toprule
    \textbf{Method} & \textbf{Avg. Collisions\,$\downarrow$} & \textbf{Success Rate (N/10)} \\
    \midrule
    \midrule
    DEM-Trav  & 6.6  & 4/10 \\
    LeSTA     & \cellcolor{second}4.6  & \cellcolor{second}6/10 \\
    \textbf{Ours}      & \cellcolor{best}0.2 & \cellcolor{best}10/10 \\
    \bottomrule
    \bottomrule
  \end{tabular}
  }
  \vspace{-0.3cm}
\end{table}

\subsection{Traversability-aware Navigation}\label{experi_2:nav}

\subsubsection{Experimental Setup}
We evaluated off-road navigation in a Gazebo simulation environment including complex hillside terrain, varying slopes, rigid obstacles (e.g., trees, rocks, and high bushes), and passable vegetation such as low bushes. The experiments employed a Husky robot equipped with an OS1-64 LiDAR. The navigation stack utilized the DWA planner~\cite{fox2002dynamic} operating on local traversability maps without a global map. Local costmaps were generated using a 0.5 threshold to distinguish occupied from passable regions, with a grid size of $8 \times 8$ m and 0.15 m resolution. For evaluation, we measured the average number of collisions and the success rate over 10 trials. After each collision, the robot executed a recovery procedure and resumed navigation. A mission was deemed unsuccessful if the recovery procedure failed.

To ensure a fair comparison, both learning-based methods (LeSTA and our approach) were trained on datasets collected from identical trajectories within the hill world environment~\cite{sanchez2022automatically}, where positive samples consisted of traversals across \textit{uneven roads}, \textit{slopes}, and \textit{low bushes}. The collected data were split into 80\% for training and 20\% for validation. The training parameters for our method and LeSTA followed the same configurations detailed in Sec.~\ref{experi_2:mapping}. For DEM-Trav, platform-specific thresholds were configured for the Husky robot, set to 0.3 m, 0.1 m, and 1.0 rad for step height, roughness, and slope, respectively.

\subsubsection{Navigation Results}
\begin{figure}[t!]
% \vspace{-0.50cm}
    \centering
        \def\width{0.5\textwidth}
        {
            \includegraphics[width=\width, clip, trim=0mm 0mm 0mm 0mm]{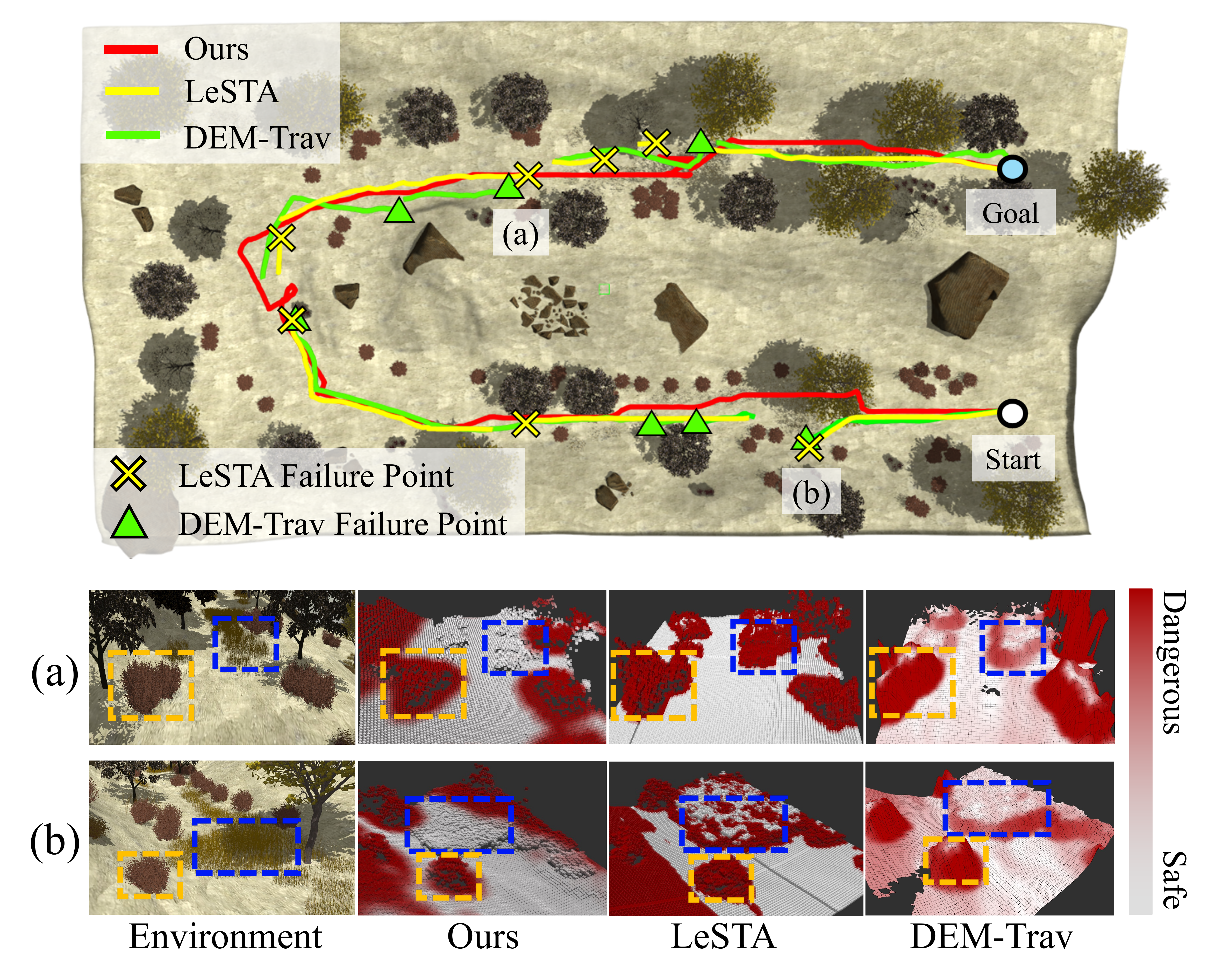}
    }
    \vspace{-0.6cm}
\caption{A representative example summarizing a successful navigation task in the simulation. Trajectory lines and collision markers are shown for each method. Detailed views (a) and (b) highlight vegetation regions, where orange boxes indicate non-traversable areas and blue boxes denote traversable bushes.}
    \vspace{-0.4cm}
    \label{fig:sim_result}
\end{figure}

As shown in \tabref{tab:mission_overall}, our method consistently reached the goal with stable trajectories, achieving a perfect success rate (10/10) and a minimal average collision count of 0.2. In contrast, both LeSTA and DEM-Trav frequently misclassified \textit{low bushes} as non-traversable, leading to multiple navigation failures with success rates of 6/10 and 4/10, respectively. The representative navigation trajectories for each method are visualized in \figref{fig:sim_result}.

Specifically, DEM-Trav relied on rigid, human-defined parameters, leading to inconsistent traversability predictions across varying slopes and step heights. Similarly, LeSTA struggled to achieve a balanced representation of diverse terrain features, which caused the misclassification of ambiguous regions, such as \textit{low bushes}. 
Consequently, the planner treated passable vegetation as obstacles, leading to navigation failures. In contrast, our method successfully navigated the complex environment by leveraging traversal experience.

\section{Conclusion \& Discussion} \label{Conclusion}

This paper presents an experience-aware framework for self-supervised traversability estimation that leverages robot traversal experience. By incorporating anomaly detection into the traversability learning, our approach overcomes the inherent limitations of \ac{PU} learning caused by the positive-only constraint. In addition, targeted augmentation strategies were applied to expand the diversity of traversal experiences, improving adaptability to unseen regions. We further validated the practical applicability of our approach through navigation tasks.

Despite its effectiveness, our framework has two limitations. First, providing supervision to empty cells may lead to training instability. Second, robot-state information such as proprioceptive signals (e.g., battery status or motor conditions) is not yet incorporated, which would be essential for a more comprehensive traversability assessment.

Future work will focus on incorporating uncertainty-aware learning to mitigate instability from incomplete measurements caused by occlusion and integrating robot-state information to achieve more robot-specific traversability estimation. In addition, as existing class-based datasets do not adequately reflect actual traversability, we plan to develop comprehensive datasets with ground-truth traversability to enable more accurate evaluation.

% \section{Acknowledgements}

\scriptsize
\bibliographystyle{Packages/IEEEtranN} % not IEEEtran, but IEEEtranN for using cite author
\bibliography{Packages/string-short, Packages/references}

% that's all folks
\end{document}